\documentclass[runningheads]{llncs}
\usepackage{booktabs}
\usepackage{makecell}
\usepackage{graphicx}
\usepackage{amsmath,amssymb,amsfonts}
\usepackage{array}
\usepackage{subfig}
\usepackage{xcolor}
\usepackage{color}
\usepackage{soul}
\usepackage{makecell}
\newcolumntype{P}[1]{>{\centering\arraybackslash}p{#1}}
\def\BibTeX{{\rm B\kern-.05em{\sc i\kern-.025em b}\kern-.08em
    T\kern-.1667em\lower.7ex\hbox{E}\kern-.125emX}}
    
\begin{document}
\title{Learning to Select the Relevant History Turns in Conversational Question Answering}
\titlerunning{Learning to Select the Relevant History Turns in ConvQA}
\author{
Munazza Zaib\inst{1}\thanks{\emph{Corresponding Author: munazza-zaib.ghori@hdr.mq.edu.au}} \and Wei Emma Zhang\inst{2} \and Quan Z. Sheng\inst{1} \and Subhash Sagar\inst{1} \and Adnan Mahmood\inst{1} \and Yang Zhang\inst{1}\\}
\authorrunning{M. Zaib et al.}

\institute{School of Computing, Macquarie University, Sydney, NSW 2109, Australia \and School of Computer and Mathematical Science, The University of Adelaide, Adelaide, Australia}

%

%
\maketitle              
\vspace{-2em}
\begin{abstract}
 The increasing demand for web-based digital assistants has given a rapid rise in the interest of the Information Retrieval (IR) community towards the field of conversational question answering (ConvQA). However, one of the critical aspects of ConvQA is the effective selection of conversational history turns to answer the question at hand. The dependency between relevant history selection and correct answer prediction is an intriguing but under-explored area. The selected relevant context can better guide the system so as to where exactly in the passage to look for an answer. Irrelevant context, on the other hand, brings noise to the system, thereby resulting in a decline in the model's performance. In this paper, we propose a framework, DHS-ConvQA (\textbf{\textit{D}}ynamic \textbf{\textit{H}}istory \textbf{\textit{S}}election in \textbf{\textit{C}}onversational \textbf{\textit{Q}}uestion \textbf{\textit{A}}nswering), that first generates the context and question entities for all the history turns, which are then pruned on the basis of similarity they share in common with the question at hand. We also propose an attention-based mechanism to re-rank the pruned terms based on their calculated weights of how useful they are in answering the question. In the end, we further aid the model by highlighting the terms in the re-ranked conversational history using a binary classification task and keeping the useful terms (predicted as 1) and ignoring the irrelevant terms (predicted as 0). We demonstrate the efficacy of our proposed framework with extensive experimental results on CANARD and QuAC -- the two popularly utilized datasets in ConvQA. We demonstrate that selecting relevant turns works better than rewriting the original question. We also investigate how adding the irrelevant history turns negatively impacts the model's performance and discuss the research challenges that demand more attention from the IR community.

\end{abstract}

\begin{table}[!h]\centering
\caption{An example of information-seeking conversation. The relevant terms in the conversational history are shown in boldface.}
{\renewcommand{\arraystretch}{1.15}}
\begin{tabular}{p{1.5cm}p{5.5cm}}

      \hline
    \hline
     \multicolumn{2}{c}{Topic: Jal-The band} \\
     \hline
    \textbf{ID}  & \textbf{Conversation} \\
    \hline
    Q1  & Who founded \textbf{Jal}? \\
    A1  & Goher Mumtaz and Atif Aslam.\\\hline
    Q2  & Where was Atif Aslam born? \\
    A2  & Wazirabad\\\hline
    Q3 & When was the \textbf{band} founded?\\
    A3 & 2002. \\ \hline
    Q4  & What was their first \textbf{album}? \\
    A4  & Aadat. \\ \hline
    Q5 & When was \textbf{it} released?  \\
    \hline
    
    \hline
    \end{tabular} 
      \label{jal}
      \vspace{-2.0em}
\end{table}
\section{Introduction}
The long-standing objective of the IR community has been to design intelligent agents, whether web-based or mobile-based, that can engage in eloquent interaction with humans iteratively \cite{DBLP:conf/acl/LiGGC22,kotov2010towards,zaib2022conversational}. The IR community has come closer to the realization of the dream owing to the rapid progress in conversational datasets and pre-trained language models \cite{zaib2022conversational}. These advancements have resulted in the birth of the field of conversational question answering (ConvQA). ConvQA provides a simplified but strong setting for conversational search \cite{DBLP:conf/sigir/Qu0QCZI19} where the user initiates the conversation with a specific information need in mind. The system attempts to find relevant information pertinent to the question at hand iteratively based on a user's response or follow-up questions \cite{DBLP:conf/sigir/Qu0QCZI19,DBLP:conf/acsw/ZaibSZ20,qu2019attentive}. When answering the follow-up questions, the model needs to take the previous conversational turns into account to comprehend the context \cite{reddy2019coqa,choi-etal-2018-quac}. Selecting the relevant context that helps the model in building a clear and strong understanding of the current question is, therefore, a very critical challenge in ConvQA \cite{qu2019attentive,DBLP:conf/aaai/QiuHCJQ0HZ21,zaib2021bert}. Adding the entire conversational history may bring the noise to the system with irrelevant context. This hinders the model's capability to correctly interpret the context of the conversation \cite{DBLP:conf/sigir/SauchukTHTS22,zaib2021bert}, thus resulting in a decline in the accuracy of the predicted answer.

\textbf{Limitations of state-of-the-art.} The process of selecting the relevant conversational turns and predicting the correct answer span is based on a number of factors. The flow of conversation keeps on changing because of the presence of dialog features like dialog shift, topic return, drill down, and clarification \cite{yatskar-2019-qualitative}. Therefore, prepending \textit{k} immediate turns, as suggested in \cite{DBLP:conf/sigir/Qu0QCZI19,DBLP:journals/corr/abs-1812-03593,DBLP:conf/iclr/SeoKFH17,ohsugi-etal-2019-simple}, won't be able to capture the gist of what the current question is about. Table~\ref{jal} shows an example of a conversational excerpt. Q2 shows a \textit{topic shift}, whereas, Q3 represents \textit{topic return}. Q4 and Q5 are examples of topic drill. The topic of Q4 is related to the band. Adding Q2, which 
inquires 
about the singer, to it would introduce noise within the input. 

Another factor is of incomplete or vague follow-up questions that impede the model from fully interpreting the conversation to be able to select the relevant conversational turns. The literature \cite{raposo2022question,kim-etal-2021-learn,DBLP:conf/wsdm/VakulenkoLTA21,yu2020few} suggests the task of question rewriting (QR) to address the issue where QR refers to rewriting the current question by adding missing information pieces or resolving co-references, thereby, making it context-independent \cite{raposo2022question}. However, taking questions out of the conversational context results in losing important cues from the conversational flow. Also, the rewritten questions might be lengthy and verbose which, in turn, adds difficulty in selecting relevant conversational history \cite{10.1145/3477495.3531815}. The model requires the resolution of `it' and information about missing context (i.e., the band) to extract the correct answer span of Q4 from Table~\ref{jal}.

\textbf{Approach.} We study and propose a framework, DHS-ConvQA (\textbf{\textit{D}}ynamic \textbf{\textit{H}}istory \textbf{\textit{S}}election in \textbf{\textit{C}}onversational \textbf{\textit{Q}}uestion \textbf{\textit{A}}nswering), that focuses on selecting the relevant conversational turns by ensuring the changing conversational flow and incomplete information requirement expressed in the query in view. The model first generates the context entities and question entities for the entire conversational history using distant supervision learning. The \textit{context entity} refers to the entity mentioned from the conversational context whereas \textit{question entity} is the entity targeted in the current question. Once the entities are generated, the turns containing non-similar context entities and question entities as compared to the current question are pruned. The remaining conversational turns are then re-ranked on the basis of their relevance to the current question. Their relevance is measured via the weightage assigned to them using the history attention mechanism. In the end, to further aid the answer prediction process, we utilize a binary classification task to highlight the key terms within the conversation history as 1 and 0. This particularly helps the model with the incomplete questions by providing hints about what the current question is about. We also compare our proposed framework to the standard question rewriting module to evaluate its effectiveness.

\textbf{Contributions:}
Our main contributions 
are as the following:
\begin{itemize}
    \item We utilize a distant supervision approach to generate context and question entities for conversational turns. The turns that do not share similar context and question entities to the current question are pruned. The remaining turns are then re-ranked on the basis of their relevance to the question. 
    \item We use binary term classification to highlight the important information from the conversational history. This helps in adding the missing information to the current incomplete question so that the model gets a better picture of the conversational flow.
    \item We demonstrate by our experimental setup that the dynamic history selection works better than question rewriting and that the presence of negative samples or irrelevant turns results in a decline in the model's performance. We conclude our paper with two possible research challenges for the IR community. 
    
\end{itemize}
\section{Related Work}

\subsection{Conversational Question Answering}
The field of ConvQA has seen a rapid boom in terms of research works and development over the past few years mainly because of the increasing demands for digital assistants \cite{zaib2022conversational,DBLP:journals/ftir/GaoGL19}. This development is further supported by the introduction of pre-trained language models and two large-scale ConvQA datasets, i.e., CoQA \cite{reddy2019coqa} and QuaC \cite{choi-etal-2018-quac}  resulting in many state-of-the-art ConQA models \cite{DBLP:conf/sigir/Qu0QCZI19,qu2019attentive,zaib2021bert,DBLP:journals/corr/abs-1812-03593,DBLP:conf/iclr/HuangCY19,DBLP:conf/ijcai/0022WZ20,DBLP:conf/acl-mrqa/YehC19,qu2020open,DBLP:conf/ecir/QuYCCKI21}. The task of ConQA can be utilized in three settings; extractive \cite{choi-etal-2018-quac,reddy2019coqa}, retrieval \cite{dalton2021cast}, and knowledge graph-based QA \cite{DBLP:conf/cikm/ChristmannRASW19,DBLP:conf/aaai/SahaPKSC18,DBLP:conf/nips/GuoTDZY18,DBLP:conf/emnlp/ShenGQGTDLJ19,kacupaj-etal-2021-conversational}. We focus on extractive ConvQA in our paper. The input to any ConvQA model generally comprises a context passage, conversational history, and current question. The way the model selects and represents conversational history has a direct impact on the prediction of the correct answer span. This presents another research challenge for the IR community.

\subsection{History Selection in ConvQA}
ConvQA is still in its infancy and has a number of critical challenges that demand attention. One such challenge is the selection of the relevant history turns and how 
to utilize them within the framework \cite{qu2019attentive,zaib2021bert}. The approaches within extractive ConvQA utilize static and dynamic methods to represent conversational history. In the case of the static history representation methods, the widely used approach involves prepending \textit{k} history turns to the current question \cite{reddy2019coqa,choi-etal-2018-quac,DBLP:conf/iclr/SeoKFH17}. On the contrary, the dynamic selection can be further categorized as \textit{hard history selection} and \textit{soft history selection.} \textit{Hard history selection} is a mechanism to select a subset of question-relevant conversational turns \cite{DBLP:conf/sigir/Qu0QCZI19,DBLP:conf/aaai/QiuHCJQ0HZ21,zaib2021bert}. However, the more pervasive and reliable method is to generate question-aware contextualized representations of the conversational history \cite{DBLP:conf/iclr/HuangCY19,qu2019attentive}. The contextualized representations are, then, passed on to the neural reader to look for the answer span within the given context passage. 

We utilize a combination of the two techniques (i.e., \textit{soft and hard history selection}) to filter out the irrelevant turns and utilize only the relevant conversational history within the model.

\subsection{Question Rewriting}
A popular research direction that aims to address the challenges pertinent to an incomplete or ambiguous question is question rewriting (QR). The task of QR is recently adopted in the field of ConvQA to reformulate the ambiguous and incomplete questions, that relies on the conversational context for their interpretation and generate self-contained questions that can be answered from the given context \cite{raposo2022question,kim-etal-2021-learn,DBLP:conf/wsdm/VakulenkoLTA21,elgohary-etal-2019-unpack,li-etal-2022-ditch,chen2022reinforced,ishii-etal-2022-integrating}. However, the task of QR takes the conversational questions out of the context by transforming them into self-contained questions which does not fit well with the whole idea of ConvQA setting \cite{DBLP:conf/sigir/ChristmannRW22}.

\section{Methodology}

\subsection{Task Formulation}
We take the traditional setting on ConvQA into consideration wherein a user instigates the conversation with a specific information need and the system attempts to provide a relevant and accurate answer after each of the user's questions \cite{zaib2021bert}. To answer each question, the model needs to refer to the previous conversation turns to get the quintessence of the context of the conversation\cite{zaib2021bert,DBLP:conf/aaai/QiuHCJQ0HZ21,DBLP:conf/sigir/Qu0QCZI19}. However, not all the previous turns contribute to aiding the model in understanding the current question. Thus, our model follows a four-step process to make sure that the most relevant terms are selected from the entire conversation and those selected turns maximize the probability of correct answer prediction by providing additional cues to the answer prediction module.

\begin{figure}[t!]
\centering
\subfloat[Pipeline approach \label{qr}]{\includegraphics[width=0.8\linewidth]{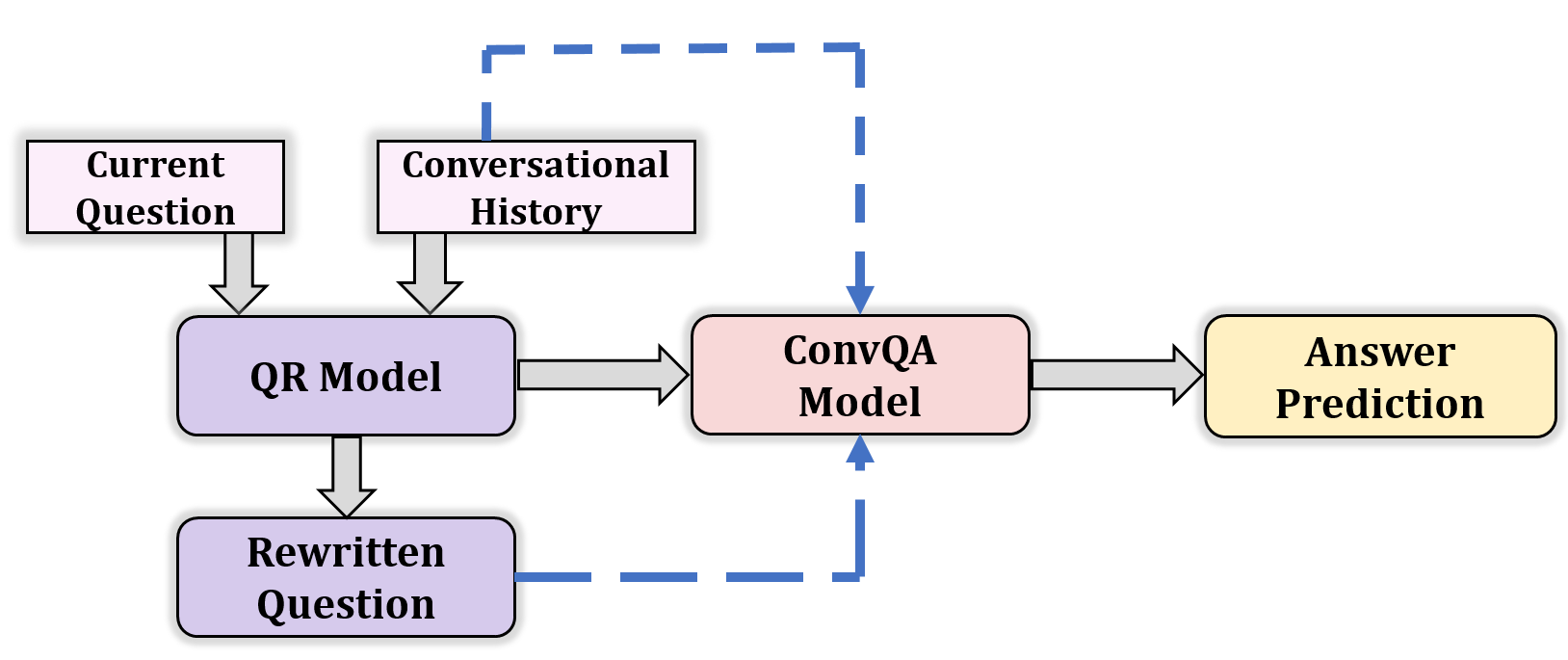}} \qquad 
\subfloat[Our proposed approach \label{op}]{\includegraphics[width=\linewidth]{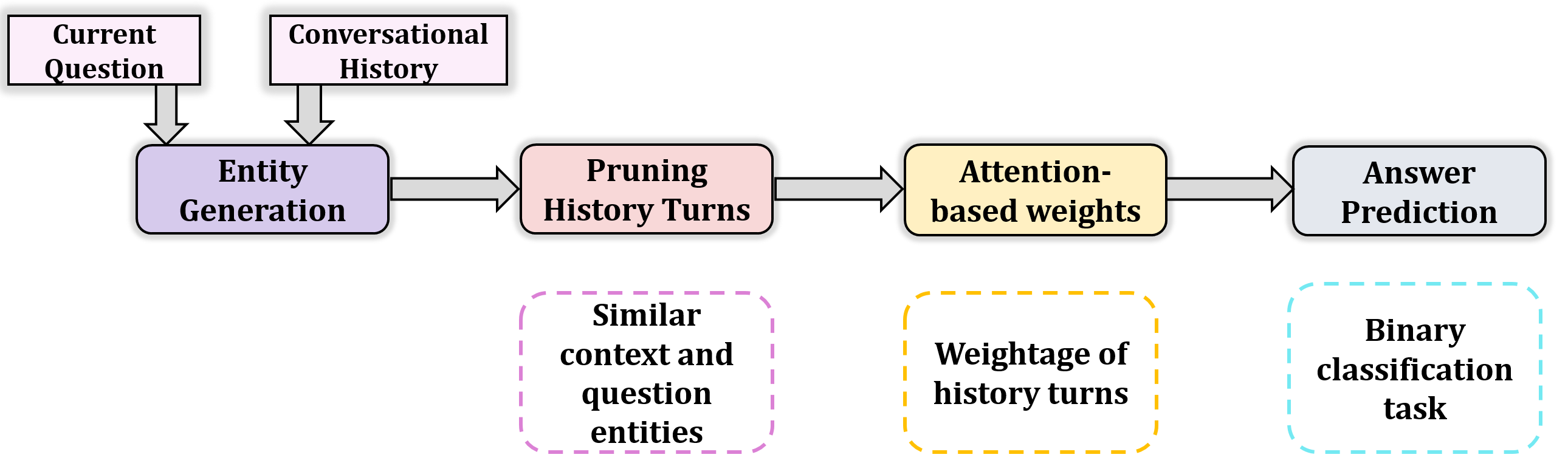}} 
\caption{In the traditional pipeline approach, a context-independent question rewrite is generated by the QR module which is then answered by the QA module. The illustration of our proposed framework shows the dotted line modules which aid the respective process and help the model in finding relevant conversational turns that can help predict the correct answer span.}
\vspace{-2.5em}
\end{figure}

More formally, given a context passage \textit{C}, current question $Q_i$, previous history turns \textit{H}, the task of our proposed framework is to select the most relevant history turns $H^{'}$ based on different factors such as having similar context and question entities to the question at hand and the order based on their weightage which shows their relevance to the question. Once the conversational turns are selected, we further aid the answer prediction process by highlighting the relevant terms using the binary classification task. 

\subsection{Pipeline Approach}
One of the most common techniques that have been in use to tackle the complexity of ConvQA tasks is by decomposing it into two sub-tasks of QR and QA \cite{kim-etal-2021-learn,vakulenko2021question,raposo2022question,vakulenko-etal-2020-wrong}. The output from the QR module serves as an input to the QA module. The QR module is responsible for re-generating the question from scratch based on the provided context and the question at hand. Different techniques are in practice to produce these rewrites such as neural networks \cite{vakulenko2021question} and pre-trained language models \cite{raposo2022question,vakulenko-etal-2020-wrong,DBLP:journals/corr/abs-2004-01909}.

The QR module can be trained on CANARD \cite{elgohary-etal-2019-unpack} dataset which consists of context-independent rewrites of the QuAC \cite{choi-etal-2018-quac} dataset. The dataset contains 40K question-answer pairs produced by human annotators. Similar to \cite{DBLP:journals/corr/abs-2004-01909}, we utilize GPT-2 \cite{radford2018improving} to train the QR module. The conversational turns and the current question are passed on as input to the module during the training process and the module is required to generate a question rewrite that is to be answered by the QA module. Since it is assumed that all the dependencies and co-references have been resolved when rewriting the question, we use a traditional QA model instead of a ConvQA model to answer the question. We put together the process of predicting an answer as follows:
\begin{equation}
  P(a_i \mid q_i, C, H) \approx P^{qa}(a_i \mid P^{qr}({q'}_i \mid qi, H), C)
\end{equation}
where $P^{qa}$ and $P^{qr}$ represent the probability of the two sub-task modules, respectively. $q^{'}_i$ represents the rewritten question by the QR module and will be provided as input to the QA module along with conversational history as shown in Figure~\ref{qr}.

\subsection{DHS-ConvQA}
The task of ConvQA heavily relies on conversational history. The more relevant and curated the conversational history is presented as input, the higher the chances of predicting the correct answer span. In our proposed method, we focus on utilizing different techniques to select the most relevant conversational turns to help the model better understand the question at hand. We emphasize addressing two issues. The first is to identify the relevance of turns to the current question. For this, we aim to generate context and question entities for each turn. To capture these entities for any incomplete question, we use the context and question entities from the last question. The underlying intuition is that the incomplete questions are usually the continuation of the conversation flow and it is safe to take the information from it to fill in the missing pieces. The context entity of Q4 in Table~\ref{jal} is \textbf{band} and the question entity is \textbf{album}. These two can easily be added to incomplete Q5 and the resultant question would be \textbf{`When was the band's album released?'} To generate these entities, we employ a seq2seq pre-trained language model, BART \cite{DBLP:conf/acl/LewisLGGMLSZ20}. The model takes the current question and the conversational history as input and is best utilized when the information is duplicated from the input but manipulated to produce the result \cite{DBLP:conf/acl/LewisLGGMLSZ20}. Once the entities are generated, the next step is to prune all the history turns where there is no similarity of context and question entities with the current question. This comes under \textit{hard history selection}.

Once the irrelevant turns are pruned, the next step is to calculate the attention weights for the remaining turns using the attention module. The attention module consists of a single-layer feed-forward network that learns an attention vector to map a sentence representation to a logit. Subsequently, the softmax probability function is utilized to calculate the probabilities across all the sequences. More formally, the computation of the weights can be shown as follows: 
\begin{equation}
    w_{i} = \frac{e^{D\cdot s_{i}^{k}}}{\sum_{i^{'}=1}^{I} e^{D\cdot s_{i}^{k^{'}}}}
\end{equation} 
where, \textit{D} is an attention vector, $s_{i}^{k}$ is a sentence representation, and $w_{i}$ is the attention weight for $s_{i}^{k}$. 

Once the weights are calculated, the vectorized turns are then passed on to the next module in a sequence where the turn with the highest weight is added next to the current question. This is how \textit{soft history selection} is utilized within the framework. The output of the attention module is then passed on to RoBERTa \cite{DBLP:journals/corr/abs-1907-11692} as an input. The next step is to introduce a term classification layer on top of the representations of the sub-token of each representation. The layer consists of a linear layer, a sigmoid function, and a dropout layer, and  outputs a scalar value for each token. The terms relevant to the current question are highlighted as \textit{`1'} and the remaining terms are set as \textit{`0'}. The terms represented as \textit{1} serve as a piece of missing information for incomplete questions. Finally, the decoder will generate the answer span for the current question based on context passage, conversational history, and the additional cues added to it.

\subsection{Training of the Model}
For the training of the entity generation module and binary term classification, we follow the strategy of distantly supervised labeling introduced in \cite{10.1145/3477495.3531815}. The idea behind the strategy is that if a piece of information is necessary for interpreting and answering the current question, it should be considered part of the current question. We start with the first question and gather all the context and question entities from it. For the incomplete or
ambiguous follow-up questions, we keep on adding these
entities to fill in the missing information. The entities are
considered to be relevant for the incomplete question if an
answer span is retrieved by adding them. For binary term classification, the relevant terms are tagged as \textit{1} for being relevant and \textit{0} for being irrelevant after passing through the term classification layer. For the task of answer prediction, the model is trained on the QuAC \cite{choi-etal-2018-quac} dataset.
\subsection{Configurations}
We train the pipeline model on around 31K pairs of original questions and their respective rewrites, and validate it on a development set of 3K question pairs. The test set of the QR task consists of 5K question pairs from the CANARD \cite{elgohary-etal-2019-unpack} dataset. The DHS-ConvQA model is trained, validated, and tested on around 100K question-answer pairs from QuAC \cite{choi-etal-2018-quac} dataset. The entire code was written in Python, making use of the popular PyTorch library \footnote{https://pytorch.org/}.

\textbf{Structured representations:} For generating the question and context entities, we utilize BART. The default hyperparameters were used from the Hugging Face library \footnote{https://huggingface.co/facebook/bart-base}. Early stopping was enabled with a batch size of 4. Adam optimizer with a learning rate of 0.00005 is used with a weight decay of 0.01.

\textbf{Binary term classification:} We utilize RoBERTa's PyTorch implementation by Hugging Face library\footnote{https://huggingface.co/docs/transformers/model\_ doc/roberta} and introduce a term classification layer on top of it. Adam optimizer is used with a learning rate set in the range of \{2e-5, 3e-5, 3e-6\}. The dropout on the term classification layer lies in the range \{0.1, 0.2, 0.3, 0.4\}. The maximum answer length is set to 40 and the maximum question length is set to 64.

\subsection{Dataset}

\subsubsection{CANARD:}
We utilize CANRAD \cite{elgohary-etal-2019-unpack} dataset to train the QR module to generate the rewrites of the given question. The CANARD dataset is based on QuAC and yields the same answer as the original questions. We use the training and development sets to train and validate the QR model and the test set to evaluate the QA module.
\subsubsection{QuAC:}
We have experimented with one of the widely utilized datasets that support the ConvQA setting--QuAC \cite{choi-etal-2018-quac}. It comprises 100K question-answer pairs in a teacher-student setting. The reason for selecting this dataset is that it embodies more dialog features than the other datasets as proved in \cite{yatskar-2019-qualitative} and gives us more scope to experiment with our relevant history selection model.

\subsection{Competing Methods}
Since our proposed method is a combination of steps, we have, therefore, selected the models that more or less follow the part of our proposed technique in their models for a fair comparison.  The chosen models are widely utilized and have been proven to perform remarkably well in ConvQA settings. These methods include:
\begin{itemize}
    \item \textbf{BERT-HAE \cite{DBLP:conf/sigir/Qu0QCZI19}:}  The BERT-based model incorporates the conversational turns with history answer embedding (HAE) to predict the correct answer span. They experimented with different conversational turn settings and found optimal answers by including 5-6 history turns.
    \item \textbf{BERT-HAM \cite{qu2019attentive}:} BERT-based history answer modeling (HAM) performs \textit{soft selection} on the relevant conversational turns. The model conducts attentive history selection based on weights assigned to them. These weights signify how relevant the turn is in answering the current question.
    
    \item \textbf{BERT-CoQAC \cite{zaib2021bert}:} Instead of prepending all the conversational turns to the current question, this model utilizes cosine similarity to select the relevant turns. 
    
    \item \textbf{CONVSR \cite{zaib2023keeping}:} The model generates the intermediate structured representations to help the model in understanding the current question better. 
\end{itemize}

\subsection{Evaluation Metrics}
For the sake of the evaluation, we follow the metrics suggested in \cite{choi-etal-2018-quac} to assess our proposed model's performance. The metrics include not only the F1 score to evaluate the accuracy of the predicted answer but also the human equivalence score for questions (HEQ-Q) and human equivalence score for dialog (HEQ-D). HEQ-Q measures the model's ability to retrieve a more accurate (or, at least, similar) answer to the current question than the humans. HEQ-D represents the same performance measure, but instead of a question, it assesses the quality of the overall conversation. 
\section{Experimentation Results and Analysis}
\begin{table*}[t]
    \caption{Performance evaluation of the pipeline approach and our model using the QuAC and CANARD datasets. The best scores are highlighted in bold.}
    \begin{center}

    \begin{tabular}{P{2.0cm}|P{2.5cm}|P{2.5cm}|P{2.5cm}|P{2.5cm}} \hline \hline

\multicolumn{1}{c|}{\textbf{Models}} & \multicolumn{1}{c|}{\textbf{Approach}} & \multicolumn{1}{c|}{\textbf{F1}} &  \multicolumn{1}{c|}{\textbf{HEQ-Q}} &  \multicolumn{1}{c}{\textbf{HEQ-D}}\\ 
   \hline \hline
    BERT-HAE & \begin{tabular}{l}Pipeline \\Ours\end{tabular} & \begin{tabular}{l}62.3 \\\textbf{63.1} \textcolor{blue}{\textbf{(+0.8)}}\end{tabular} & \begin{tabular}{l}58.2 \\\textbf{58.9} \textcolor{blue}{\textbf{(+0.7)}}\end{tabular}& \begin{tabular}{l}5.5 \\\textbf{6.0} \textcolor{blue}{\textbf{(+0.5)}} \end{tabular} \\
    \hline
   BERT-HAM  & \begin{tabular}{l}Pipeline \\Ours\end{tabular} & \begin{tabular}{l}63.4 \\\textbf{65.4} \textcolor{blue}{\textbf{(+2.0)}}\end{tabular} & \begin{tabular}{l}60.1 \\\textbf{61.8} \textcolor{blue}{\textbf{(+1.7)}}\end{tabular} & \begin{tabular}{l}6.1 \\\textbf{6.7} \textcolor{blue}{\textbf{(+0.6)}}\end{tabular}\\ 
   \hline
    BERT-CoQAC  & \begin{tabular}{l}Pipeline \\Ours\end{tabular} & \begin{tabular}{l}63.1 \\\textbf{64.4} \textcolor{blue}{\textbf{(+1.3)}}\end{tabular}&\begin{tabular}{l}59.2\\\textbf{59.9} \textcolor{blue}{\textbf{(+0.7)}}\end{tabular} & \begin{tabular}{l}5.9\\\textbf{6.9} \textcolor{blue}{\textbf{(+2.0)}}\end{tabular}\\
    \hline
    CONVSR  & \begin{tabular}{l}Pipeline \\Ours\end{tabular} & \begin{tabular}{l}66.1 \\\textbf{67.5} \textcolor{blue}{\textbf{(+1.4)}}\end{tabular}&\begin{tabular}{l}62.2\\\textbf{65.3} \textcolor{blue}{\textbf{(+3.1)}}\end{tabular} & \begin{tabular}{l}6.0\\\textbf{7.5} \textcolor{blue}{\textbf{(+1.5)}}\end{tabular}\\
    \hline

    \end{tabular}
    \end{center}
    \label{result12}
    \vspace{-2em}
    
\end{table*}

We conduct experiments on our proposed model using the QuAC \cite{choi-etal-2018-quac} and CANARD \cite{elgohary-etal-2019-unpack}
datasets and compare the results with the competing models.
\subsection{\textbf{DHS-ConvQA is Viable for Selecting Relevant Conversational Turns}}
Topic shift and topic return are two main challenges in the field of ConvQA. Adding \textit{k} immediate turns as a part of the input to the ConvQA model fails to capture the essence of the conversational flow. Also, rewriting a question takes it out of the conversational context and focuses more on generating high-quality rewrites instead of improving the performance of a ConvQA model. Thus, the first and foremost takeaway from our experimental results is that selecting relevant history turns aids the model in better understanding the question at hand and then predicting the accurate answer span. Instead of rewriting the questions to fill in the missing gaps, which takes out the questions from the conversational context, selecting relevant turns after going through different stages works well in yielding higher accuracy as shown in Table~\ref{result12}.
\subsection{\textbf{Role of Relevant Conversational History in the ConvQA Setting}}

The existing works either opt for \textit{soft history selection} or employ \textit{hard history selection}. We propose a combination of both along with highlighting relevant terms to the current question as additional cues. From Table~\ref{result1}, we can clearly deduce that our proposed 
model consistently improves the model's performance, thereby,
confirming the fact that our model works well in the ConvQA setting. We also conduct an in-depth analysis of the proposed model by studying the effect of each module it brings within the framework. Table~\ref{result1} shows that omitting the pruning of the turns step results in a greater decline in the F1 score as compared to the other modules. The underlying reason is that without pruning, the model considers all the conversational turns as a part of input which brings in the noise in terms of irrelevant turns.

\begin{table}[tb!]
\caption{The evaluation results of our proposed model with the competing methods on the QuAC dataset. We also demonstrate the effect of each module on the model's performance.}
\begin{center}
{
\begin{tabular}{P{4.0cm}|P{2.5cm}|P{2.5cm}|P{2.5cm}} \hline
    \hline
\multicolumn{1}{c|}{\textbf{Model}} & \multicolumn{1}{c}{\textbf{F1}} &  \multicolumn{1}{c|}{\textbf{HEQ-Q}} &  \multicolumn{1}{c}{\textbf{HEQ-D}} \\
\hline
   \hline
    BERT-HAE & 63.1  & 58.9 & 6.0   \\
\hline
   BERT-HAM & 65.4  & 61.8 & 6.7  \\ 
\hline  
BERT-CoQAC & 64.4 & 59.9 &\textbf{6.9} \\
\hline
Ours (w/o pruning) & 64.3 & 62.9 & 6.6 \\ \hline
Ours (w/o re-ranking) & 67.3 & 63.6 & 6.9 \\ \hline
\begin{tabular}[c]{@{}c@{}}Ours (w/o term\\  classification)\end{tabular} & 65.7 & 62.0 & 6.5\\ \hline
\textbf{Ours (complete setup)} & \textbf{67.5} & \textbf{65.3} & 7.5 \\ 
\hline
    \end{tabular}}
    \end{center}
    \label{result1}
  \vspace{-1em}
\end{table}

\subsection{\textbf{Effect of Negative Samples on Model's Performance}}
For each question, we experiment by injecting negative samples together with the relevant turns identified by the proposed model. The negative samples are the questions related to the same topic but from different passages of the QuAC dataset. They are semantically closer to the relevant questions and, therefore, they can be considered a part of conversational history by the model. From Table~\ref{result2}, we can interpret that adding negative samples results in the decline of the model's F1 score. Also, the clarification questions are least impacted by the added noise as compared to the topic return and topic shift questions. The negative samples can easily be misleading for the model to capture the gist of the changing conversational flow.

\begin{table}[tb!]
\caption{The evaluation of the model's performance where the model receives Negative Samples (NS) as a part of input together with the relevant conversational turns.}
\begin{center}
\begin{tabular}{P{3.5cm}|P{1.0cm}|P{2.0cm}|P{2.0cm}|P{2.0cm}} \hline
    \hline
\multicolumn{1}{c}{\textbf{Model}} & \multicolumn{1}{c}{\textbf{F1}} &  \multicolumn{1}{c}{\textbf{Clarification}} &  \multicolumn{1}{c}{\textbf{Topic Shift}} &  \multicolumn{1}{c}{\textbf{Topic Return}} \\
\hline
   \hline

\begin{tabular}[c]{@{}c@{}}\textbf{Ours (complete}\\  \textbf{setup)}\end{tabular}  & \textbf{67.5} & \textbf{90.9} & \textbf{85.4} & \textbf{82.2} \\ 
\hline
    Ours    + 1NF & 67.0 & 90.4 & 82.0 & 79.5 \\ 
\hline
Ours + 3NF & 64.3 & 89.4 & 78.4 & 77.3 \\ 
\hline
Ours + 5NF & 62.5 & 88.1 & 73.5 & 73.9\\ 
\hline
Ours + 7NF & 60.7 & 86.7 & 70.9 & 70.3 \\ 
\hline
Ours + 9NF & 54.6 & 85.0 & 66.8 & 65.0 \\ 
\hline
Ours + 11NF & 52.4 & 83.8 & 63.0 & 62.1 \\ 
\hline
    \end{tabular}
    \end{center}
    \label{result2}
   \vspace{-1em}
\end{table}

\subsection{\textbf{Effect of Pruning on the Subsequent Modules In-line}}
From Table~\ref{result3}, it is clearly evident that the pruning of irrelevant conversational history has a direct effect on the performance of the rest of the modules. The better the performance of the first module, the higher the chances of correct answer prediction. If the turns are pruned accurately (100\%), the overall performance of all the components would be higher. The performance decreases as the number of correct pruned turns decreases. However, there are high chances of error propagation because the output of each module serves as an input to the next module.

\begin{table}[tb!]
\caption{Effect of pruning (in \%) on the rest of the modules }
\begin{center}
\resizebox{1\linewidth}{!}{
\begin{tabular}{P{3.5cm}|P{3.5cm}|P{3.5cm}|P{3.5cm}} \hline
    \hline
\begin{tabular}[c]{@{}c@{}}\textbf{Pruning (\%)}\end{tabular} & \textbf{Re-ranking (\%)} &  \begin{tabular}[c]{@{}c@{}}\textbf{Binary-term} \\ \textbf{Classification} (\%)\end{tabular} &  \textbf{Answer Prediction (\%)} \\
\hline
   \hline
    100 & 100  & 92 & 90   \\
\hline
   70 & 95 & 86 & 80  \\ 
   \hline  
50 & 89 & 70 & 65 \\
\hline
    \end{tabular}}
    \end{center}
    \label{result3}
   \vspace{-1em}
\end{table}

\section{Conclusion and Future Work}
This paper discusses a significant point of view on the basic concept of the role of relevance in conversational question answering (ConvQA). We argue that many existing research works, even the popular ones, do not take into account the idea of relevant history selection and modeling. We propose a framework that combines the notion of both \textit{hard history selection} and \textit{soft history selection} to curate the input for the answer prediction module carefully. The model first generates context and question entities using distant supervision learning and selects the relevant terms using \textit{`hard history selection'}. After the pruning of irrelevant terms, the model assigns attention-based weightage to the remaining turns. The assigned score is based on how relevant they are to the current question and accessed in the same order. To further aid the answering prediction process, we utilize binary classification task to highlight the important terms with respect to the current question from the conversational history. Our experimental results depict that the proposed method has the potential to change how conversational history could be utilized more effectively.

We also highlight two significant future research challenges. 
The first challenge is that 
ConvQA is essentially a modular or cascading architecture that can be categorized as an information retrieval module responsible for selecting the relevant turns and the question answering module responsible for predicting the answer span. Any negative samples or turns selected during the history selection process would directly affect the model's performance in predicting the correct answer span. Thus, there is a need for a mechanism to minimize the retrieval of irrelevant conversational turns. The second challenge centers on 
the 
development of a framework that would eliminate the chances of error propagation within the modules.


%
\bibliographystyle{unsrt}
\bibliography{acmart}
\end{document}